# Domain-Specific Deep Learning Feature Extractor for Diabetic Foot Ulcer Detection

Reza Basiri[1,2], Milos R. Popovic[1,2], Shehroz S. Khan[1,2]
1. Institute of Biomedical Engineering, University of Toronto, Toronto, Canada
2. KITE, Toronto Rehabilitation Institute, University Health Network, Toronto, Canada

*Abstract*— Diabetic Foot Ulcer (DFU) is a condition requiring constant monitoring and evaluations for treatment. DFU patient population is on the rise and will soon outpace the available health resources. Autonomous monitoring and evaluation of DFU wounds is a much-needed area in health care. In this paper, we evaluate and identify the most accurate feature extractor that is the core basis for developing a deep learning wound detection network. For the evaluation, we used mAP and F1-score on the publicly available DFU2020 dataset. A combination of UNet and EfficientNetb3 feature extractor resulted in the best evaluation among the 14 networks compared. UNet and Efficientnetb3 can be used as the classifier in the development of a comprehensive DFU domain-specific autonomous wound detection pipeline.

## I. INTRODUCTION

Diabetic Foot Ulcer (DFU) commonly occurring on the plantar foot is a type of skin wound requiring special care. Failure to heal DFUs results in long-lasting treatments with high recurrence rates, loss of mobility, independence, quality of life, amputation and even death [1], [2]. The global and North America prevalence of DFUs are reported to be 6.3% and 13% of the diabetic patient population, respectively [3].

Elderly patients are particularly burdened by DFU. The main causes of DFU are peripheral neuropathy, foot deformities and peripheral arterial disease that are exacerbated by the high rate of diagnosed diabetes among elderly patients (>65 yrs), which is approximately 6-10% [4] or about 20% when including undiagnosed diabetes [5]. Additional threat factors are poor vision, gait abnormalities, and reduced mobility [6]. The risk of major DFU-related amputations increases with age, along with the increased prevalence of these risk factors. The relative risk of amputation has been reported to be at least five times higher in older (>80 years) than in younger (40-59 years) people [6]. A substantial and increasing proportion of the elderly population is therefore at risk of complications arising from DFU.

In DFU care, where specialized health resources are outnumbered by the large growing patient population, treatment optimization is the key to sustainable health care [7]. For DFU evaluation, clinicians perform a series of common steps that are recording and examining wound locations and sizes over a period of time. Automated localization and measurement of wounds reduce manual steps in the treatment of DFU as well as facilitate potential autonomous at-home evaluation opportunities [8], [9].

For the autonomous localization of wounds, both traditional machine learning (ML) [10] and deep learning (DL) [11] methods have been used. The ML methods are limited in targeting a finite number of features in an image. These features are often not comprehensive and generalize to a subset of wounds with particular controlled environments. For example, the wound images used in several previous works using ML approaches [12], [13], generally have very few backgrounds, or their environmental backgrounds are purposefully simple or manually cropped, which could affect the generalization of these methods in real-world settings.

On the other hand, DL detection methods such as EfficientDet and Faster R-CNN have shown promising results when evaluated on wound images [14]. The DL methods' advantages are the minimal human interference and the power to learn complex wound variations. DL wound detections may differ in network dimensions, annotations, architectures and loss function strategies. But at their core, all detection DL methods rely on a limited number of classification feature extractors. However, it is unclear from previous works which feature extractor may perform best on DFU datasets.

In this study, we disregarded the auxiliary portions of these DL methods (e.g. augmentations, pre- and post-processing, thresholding steps, bounding box loss and other complimentary localization loss solutions), to find the most reliable and optimal existing core classification engine for the detection of DFUs in an image.

Our key contributions are:
- We compared 14 DL feature extractor models on four architectures including a cumulative loss function consisting of geometric-based (Dice [15]), distribution-based (Focal [16]) and distance-based (Jaccard [17]) features.
- We show that EfficientNetb3 [18] feature extractor with UNet [19] backbone deliver the best significant performance in terms of mean average precision (mAP) on the DFU dataset.

## II. BRIEF LITERATURE REVIEW

For the autonomous localization of wounds, Gholami et al. [10] compared seven traditional segmentation algorithms on a mix of wound images. The algorithms included region-based, edge-based, and texture-based methods. Based on segmentation metrics, Gholami concluded that an edge-based method that requires significant user oversight is the best clinically acceptable method.

Automatic feature learning (and thus extraction) is the main benefit of a DL model. Feature extractors are the sequence of neural network layers that are trained on datasets to substitute traditional statistical filter algorithms.

Since the inception of AlexNet [20], many DL feature extractors have been developed with architectural differences [21], [22]. The key differences that set these extractors apart are the utilizations of activation functions, pooling layers, residual connection, batch normalization techniques and dynamic network dimension. For example, EfficentNet's main contribution was scaling of width, depth and resolutions in a relative and compound format. Out of the many feature extractors, Inception [23], ResNet [24] and EfficientNet. have been consistently used with superior results in classification, object detection and semantic segmentation evaluations on MS COCO [25] and ImageNet [26] datasets. Other noteworthy feature extractors with competitive pixel-wise classification accuracy appropriate for semantic segmentation are Xception [27], and DenseNet [28]. Baseline models with relatively simple architectures are VGG [29] and MobileNet [30]. The average Top-5 percentage error range for the aforementioned extractors on ImageNet varies from 3% to 10% [31], which we decided to select as our cut-off threshold for selecting feature extractors. In this study, we included seven major feature extractors and seven variations of those, totalling 14 feature extractors.

## III. METHODS

### A. Architecture:

The most common architectures for combining high and low-level information are described in the shapes of UNet, LinkNet [32], PSPNet [33] and FPN [34]. For the most part, these architectures resemble autoencoders and vary in skip connections and layer placement orders in different arms (i.e. encoding and decoding). By combining these architectures with feature extractors, we create a computationally feasible network that consists of one final dense layer.

Two important aspects of image classification are accurate pixel class association and precise quantification of objects. Dice and Jaccard are two losses that can effectively evaluate the segmented regions. Dice, which is an inverse of the F-score, represents a weighted average of precision and recall for segmentation, while Jaccard which is an inverse of Intersect over Union (IoU) score optimizes the model for accurate pixel class assignments. As the wounds are proportionally smaller than feet, thus inherently impose class imbalance. We decided to include focal loss, which has been shown constructive in imbalanced data situations [16]. Therefore, the segmentation portion of our model was assigned an overall loss function ($\mathcal{L}_{Seg}$) consisting of Dice ($\ell_D$), Jaccard ($\ell_J$) and Binary Focal ($\ell_F$) losses as shown in equation (1). The losses' hyperparameters $(\alpha, \beta, \gamma)_{sg}$ were determined empirically by the validation set.

$$\mathcal{L}_{Seg} = \alpha_{sg}\ell_F + \beta_{sg}\ell_D + \gamma_{sg}\ell_J \quad (1)$$

### B. Feature Extractor:

A wide variety of seven commonly used feature extractors and their varieties were included. In total, we investigated 14 feature extractors as listed in Table 1. Overall, our selected architectures in combination with the feature extractors encompass the core classification engine for the majority of the object detection approaches that currently exist. One major point of significance between these approaches is the inference speed and number of parameters, which may be important for some applications but not the focus of our study. In our architectures, feature extractors were only used in the encoder arm. The decoder arm which consisted of some convolution layers remained constant. However, the weights in both arms were updated during the training. For example, in an EfficientNet-UNet combination, the encoder arm was constructed using EfficientNet and the decoder arm using six convolution layers and three skip connections as illustrated in Fig 1.

### C. Training Methodology

#### 1) Dataset:

We used 2,000 640x480 DFU RGB images from Yap et al. [35], [36] to investigate optimal architecture and feature extractors. Additionally, we incorporated 700 healthy feet images from an internet search into our dataset to reduce false positive rates. In total, the training dataset contained 2,700, 640x480 images of DFU and healthy feet. Wound quantities per image ranged from zero to five in this dataset. For this dataset, wound area occupation ranged from 0.06% to 57.38% of the entire image area.

#### 2) Pre-Processing:

Image pixel values were scaled between 0 and 1 and then zero-centred for each RGB colour channel distribution with respect to the ImageNet dataset. We performed pixel-level augmentations that include contrast, brightness, gaussian noise, blur, perspective, hue saturation and translational augmentations such as cropping, flips and rotations. Zero constant padding was done for the added borders for any translational augmentation. For training, we included 15 full 640x480 sized images per batch with a decaying learning rate from 1e-03 to 1e-08.

#### 3) Evaluation Metrics:

We used standard F1-score, IoU ratio and mAP to evaluate the training performance. For testing, we used the test-DFU2020 dataset [35], [36] which contains 2,000, 640x480 mix of DFU and healthy foot images. The ground truth for our dataset was annotated in bounding box format with min:max coordinates. For the test dataset, we did not have access to the ground truths or any control over the evaluation metric methods. Therefore we IoU is not calculated for the test dataset. The results from the test dataset were uploaded to the DFU2020 competition website [37], which returned F1 and mAP scores.

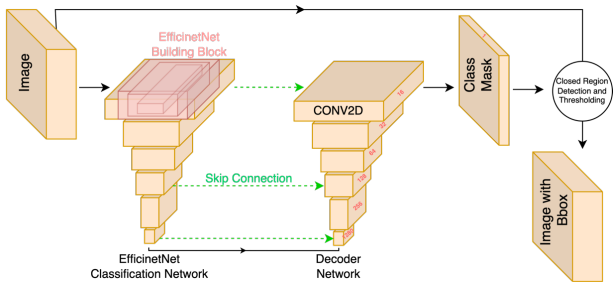

**Figure 1. A wound detection network with UNet architecture and Efficinebtb3 backbone for feature extraction.** There are three skip connections adding encoder weights to the respective decoders to maintain spatial information of layers after the flat top layer joins the two arms. Detected regions of interest are passed through a fixed minimum area and confidence level thresholding to eliminate non-wound detected regions.

*D. Inference:*

Our networks outputted wound predicted segments. In the outputs, each pixel's confidence level was between 0 and 1 via a sigmoid function. The test dataset annotations were done using bounding boxes; therefore, we added a fixed closed region detection function to translate the segmented regions to boxes. To eliminate minor and random noisy regions, a fixed minimum threshold of 0.6 and 200 pixels for mean confidence level and area were imposed to exclude noisy segments from the final bounding box coordinates.

## IV. RESULTS

To find the optimal segmentation network, we initially compared all feature extractors in combination with UNet on a small dataset that consisted of only 400 images out of the 2,000 DFU image dataset. The initial comparison was carried out using the highest validation F1-score as supported by the t-test. The results are shown in Table 1 and Fig. 2.

**Table 1. Evaluation of 14 common feature extractors with UNet architecture on the DFU2020 dataset.** F1-score was the determining factor in selecting the best-performing combinations that are denoted in bold.

| Networks | | Train Dataset | | Validation Dataset | t-test |
|---|---|---|---|---|---|
| Feature Extractor | Parameters (millions) | F1 Score | IoU Score | F1 Score | $p$ Value |
| MobileNetV2 | 8 | 0.934 | 0.877 | 0.940 | 0.886 |
| **DenseNet-121** | **12** | **0.931** | **0.872** | **0.950** | **0.015** |
| **EfficientNetb2** | **14** | **0.947** | **0.900** | **0.950** | **0.015** |
| **EfficientNetb3** | **17** | **0.939** | **0.886** | **0.951** | **0.008** |
| DenseNet-169 | 19 | 0.936 | 0.881 | 0.946 | 0.131 |
| VGG16 | 24 | 0.768 | 0.633 | 0.910 | ~0.000 |
| ResNet-34 | 24 | 0.888 | 0.803 | 0.940 | 0.886 |
| **EfficientNetb4** | **26** | **0.951** | **0.907** | **0.951** | **0.008** |
| DenseNet-201 | 26 | 0.914 | 0.844 | 0.946 | 0.132 |
| VGG19 | 29 | 0.782 | 0.652 | 0.914 | ~0.000 |
| Inceptionv3 | 30 | 0.92 | 0.854 | 0.942 | 0.669 |
| ResNeXt-50 [38] | 32 | 0.924 | 0.859 | 0.939 | 0.669 |
| ResNet-101 | 52 | 0.881 | 0.790 | 0.943 | 0.479 |
| Inception-ResnetV2 [39] | 62 | 0.939 | 0.886 | 0.945 | 0.212 |

From this comparison and while considering validation F1-scores as a distribution, the two baseline VGGs scored in the lower bound as outliers. EffcientNet variances and DenseNet-121 were upper-bound outliers with significant $p$ values based on a t-test evaluation. Furthermore, we repeated model training of the top four significantly different performing networks (EfficientNetb2, 3, and 4 and Densenet121) from this small dataset experiment. However, for retraining, we used a larger dataset including the entire 2,000 DFU image. The results from the large dataset training and evaluation on another test-DFU dataset are shown in Table 2.

**Table 2. Re-evaluating the top 4 feature extractors on UNet.** A larger test-DFU2020 dataset was used where the evaluator was blind to the ground truths.

| Networks | Train Dataset | | Test Dataset | | t-test ($p$-Value) | |
|---|---|---|---|---|---|---|
| Feature Extractors | F1 Score | IoU Score | F1 Score | mAP Score | F1 | mAP |
| EfficientNetb2 | 0.931 | 0.871 | 0.694 | 0.643 | 0.861 | 0.584 |
| **EfficientNetb3** | **0.928** | **0.867** | **0.706** | **0.658** | **0.075** | **0.040** |
| EfficientNetb4 | 0.905 | 0.834 | 0.697 | 0.641 | 0.727 | 0.426 |
| DenseNet-121 | 0.866 | 0.777 | 0.683 | - | 0.047 | - |

When training the network on 2,000 DFU images and evaluating on a separate test-DFU 2,000 images that are a mix of DFUs and healthy feet, EfficientNet variances outperformed Densenet121 in the test F1-score. Furthermore, among different variants, EfficientNetb3 significantly outperformed others in t-test evaluation on mAP. To narrow down our final segmentation network, we repeated the experiment by using the top-performing feature extractor that is EfficientNetb3 with LinkNet, PSPNet and FPN architectures as shown in Table 3.

**Table 3. Evaluation of EfficientNetb3 with different network architectures.** Combination of UNet and EfficientNetb3 consistency outperforms the other combinations when evaluated on the test-DFU2020 dataset.

| Networks | Train Dataset | | Test Dataset |
|---|---|---|---|
| Archi. | F1 Score | IoU Score | F1 Score |
| **UNet** | **0.928** | **0.867** | **0.706** |
| LinkNet | 0.919 | 0.851 | 0.683 |
| PSPNet | 0.874 | 0.779 | 0.585 |
| FPN | 0.921 | 0.856 | 0.690 |

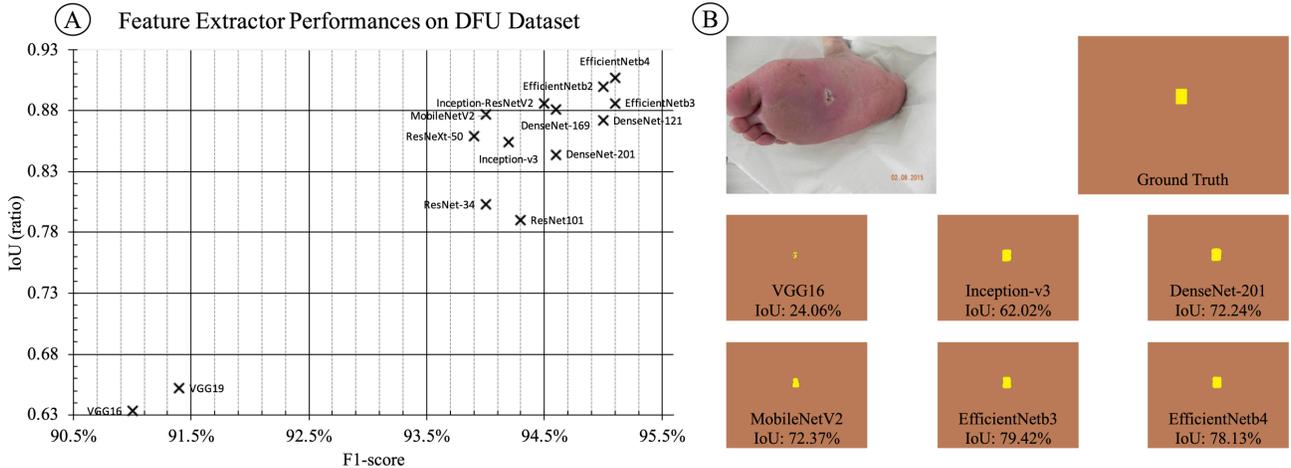

**Figure 2. Convergence of the feature extractors on UNet for DFU2020 training data and sample predicted masks.** (A) Mainly EfficientNet networks are more capable of learning fine details of DFUs based on F1 and IoU plot. (B) A sample test image, its ground truth and predicted masks from some of the investigated networks. EfficientNetb3 provides the highest IoU value for this test sample when compared to the ground truth.

Focusing on test dataset F1-scores, it is apparent that UNet and EfficientNetb3 combination is outperforming other combinations in DFU detection. The 0.706 F1-score is the highest value we achieved using the UNet+EfficientNetb3 combination without incorporating any auxiliary portions to finetune for bounding box estimations.

## V. DISCUSSION

With the rapid global increase of diabetes and the life-threatening risk factors of DFU, maintaining sustainable DFU care is facing resource allocation challenges [40]–[42]. This is especially important among the older adult patient population that is at greater risk of vascular and diabetes complications [6]. On average, due to mobility difficulties, DFUs are more common among the older patient population [>65] in comparison to the younger patient population [6]. An appropriate at-home monitor and evaluation DFU system for senior patients should be autonomous, simple and highly accurate. DL-based object detection has the potential to provide a basis for the development of such system.

Generic object detection networks such as YOLO [43] or EfficientDet are tuned to perform rapid classification and localization of a variety of objects. However, they are not optimized to perform medically acceptable segmentation of the entire object region. In fact, the recent single-shot detector advancements [44], [45] have been mainly focused on reducing computation time, and optimizing recall and mAP of the predicted classes, while compromising individual pixel classification accuracy which is an important factor for small objects. However, in the DFU domain where exact area measurement down to the pixels is important, we need a more accurate segmentation network with high IoU values. Inception-ResnetV2 has been reported as the most accurate detection network on the COCO dataset with an mAP of 35.6. While the COCO dataset compromises several generic classes and the results may not apply to the DFU dataset, we included the feature extractor from this network in our analysis for comparison.

### A. Domain-Specific Network

Our analysis highlights the importance of developing domain-specific networks. Many complex networks, such as FPN+EfficientNetb7 and Inception-ResnetV2 have been reported as the most accurate networks on generic datasets (e.g. COCO). However, we do not observe similar results in the DFU dataset using Inception-ResnetV2. We were not able to evaluate FPN+EfficientNetb7 due to the large size of this network and its incompatibility with our available hardware. EfficientNetb4 with more parameters than EfficientNetb3 did not provide any improvement in the test-DFU data evaluation. Therefore, we did not seek to evaluate the other larger networks, b5-b7 versions of EfficientNet.

### B. Our Finding

The first step to developing a comprehensive wound detection network is to identify an appropriate domain-specific feature extractor. Based on our analysis, the EfficientNetb3 feature extractor provides the highest IoU and F1 values. When trained on 2,000 images and tested on another 2,000 images, we achieved 0.706 for the F1-score. While our network was not finetuned and optimized for bounding box estimation by including box coordinate complementary regression loss functions, the results are competitive with top-performing models in the DFU2020 competition [37]. The best F1 score in this competition was 0.743 and our 0.706 was third on the leaderboard. The top performing method in this competition was a version of Faster R-CNN where convolution layers in some layers of the ResNet feature extractor were replaced by deformable convolution layers to enable deformations in the grid sampling thus increasing the accuracy in bounding box estimations. Based on our comprehensive evaluation of feature extractors and architectures, we believe an enhanced EfficientNetb3+UNet for box detection, has the potential to outperform the current top-performing DFU detection network.

### C. Limitations

There are two main limitations associated with DFU data. First, the training data size was limited to 2,000 DFU images, out of which a small portion of them consist of the wound pixels. Given a larger training dataset, the performance of the networks is expected to improve. Second, the test dataset annotations are not accessible with evaluation metrics limited to the F1-score and mAP.

## VI. CONCLUSION AND FUTURE DIRECTIONS

Autonomous DFU monitoring and evaluation could be benefitted from domain-specific DL networks. The evaluation metrics from generic datasets do not necessarily apply to DFU-specific features and applications. Efficentb3 feature extractor in combination with UNet architecture outperformed the other existing combination in the field of DFU. This proposed combination is ideal for the construction of a robust autonomous wound detection DL network. After successfully identification of an accurate DFU domain feature extractor, the next step is to construct a network with auxiliary parts to eliminate pre-set threshold values. Auto-thresholding will provide a complete autonomous platform based on confidence levels specific to the DFU domain. The auto-pixel-wise probability and auto-area constraint can be implemented in a data-driven manner by using methods such as joint learning. As part of this ongoing project, we are currently collecting DFU images in collaboration with Alberta Health Services, Canada, using thermal, depth and RGB cameras. In future we will validated the presented models on this larger dataset.